\documentclass{article}

\PassOptionsToPackage{numbers}{natbib}

\usepackage[preprint]{neurips_2024}

\usepackage[utf8]{inputenc} 
\usepackage[T1]{fontenc}    
\usepackage{hyperref}       
\usepackage{url}            
\usepackage{booktabs}       
\usepackage{amsfonts}       
\usepackage{nicefrac}       
\usepackage{microtype}      
\usepackage{xcolor}         
\usepackage{comment}
\usepackage{array}

\usepackage{amsmath,amssymb,amsfonts}
\usepackage[short]{optidef}
\usepackage{enumitem}
\usepackage{amsthm}
\usepackage{algorithm}
\usepackage{algpseudocode}
\usepackage{graphicx}
\usepackage{stfloats}
\usepackage[caption=false,font=footnotesize]{subfig}

\newtheorem{assumption}{Assumption}

\newtheorem{proposition}{Proposition}
\newtheorem{corollary}{Corollary}

\newcommand{\bdeta}{\boldsymbol{\eta}}
\newcommand{\bphi}{\boldsymbol{\phi}}
\newcommand{\bpsi}{\boldsymbol{\psi}}
\newcommand{\btheta}{\boldsymbol{\theta}}

\newcommand{\bE}{\mathbf{E}}
\newcommand{\bP}{\mathbf{P}}
\newcommand{\bL}{\mathbf{L}}
\newcommand{\bg}{\mathbf{g}}

\title{Non-Federated Multi-Task Split Learning\\ for Heterogeneous Sources}

\author{%
  Yilin Zheng, Atilla Eryilmaz \\
  Department of Electrical and Computer Engineering\\
  The Ohio State University\\
  Columbus, OH 43210 \\
  \texttt{zheng.1443, eryilmaz.2@osu.edu} \\
}

\begin{document}

\maketitle

\begin{abstract}
With the development of edge networks and mobile computing, the need to serve heterogeneous data sources at the network edge requires the design of new distributed machine learning mechanisms. As a prevalent approach, Federated Learning (FL) employs parameter-sharing and gradient-averaging between clients and a server. Despite its many favorable qualities, such as convergence and data-privacy guarantees, it is well-known that classic FL fails to address the challenge of data heterogeneity and computation heterogeneity across clients. 
Most existing works that aim to accommodate such sources of heterogeneity stay within the FL operation paradigm, with modifications to overcome the negative effect of heterogeneous data. In this work, as an alternative paradigm, we propose a Multi-Task Split Learning (MTSL) framework, which combines the advantages of Split Learning (SL) with the flexibility of distributed network architectures. In contrast to the FL counterpart, in this paradigm,  heterogeneity is not an obstacle to overcome, but a useful property to take advantage of. As such, this work aims to introduce a new architecture and methodology to perform multi-task learning for heterogeneous data sources efficiently, with the hope of encouraging the community to further explore the potential advantages we reveal. To support this promise, we first show through theoretical analysis that MTSL can achieve fast convergence by tuning the learning rate of the server and clients. Then, we compare the performance of MTSL with existing multi-task FL methods numerically on several image classification datasets to show that MTSL has advantages over FL in training speed, communication cost, 
 and robustness to heterogeneous data. 
 
\end{abstract}

\section{Introduction}

In modern edge networks, each client can have its own data source and computation limitation. Therefore, the classic Federated Learning (FL) \cite{mcmahan2017communication, kairouz2021advances} that aims to fit a common model for both the server and clients can have significant performance limitations when dealing with data and client heterogeneity \cite{zhao2018federated, zhu2021federated}. Multi-Task Learning (MTL) \cite{smith2017federated, zhang2018overview} is a natural way to evaluate a machine learning model in a heterogeneous setup, where clients can have related but different learning objectives. Existing multi-task learning methods in distributed setups mainly focus on modification of the FL framework. While having improvements in multi-task performance, these methods still keep the parameter sharing (federation) process between clients and a server.  However, this federation process may not yield the best performance in the MTL setup. In addition,  as the model size becomes larger, the FL-based methods can have high communication and computation costs.

As an alternative to FL, Split Learning (SL) \cite{gupta2018distributed, vepakomma2018split} reduces the communication cost by splitting a large model into smaller pieces. In the common-task scenario where data sources are assumed to be homogeneous, the performance of SL can fall short of FL due to imbalanced updates between clients and server \cite{joshi2021splitfed, pal2021server}. Therefore, SL is often used in combination with FL \cite{thapa2022splitfed,liao2023accelerating}. However, under a networked setup with multiple heterogeneous clients,  SL has the capability of using different models and processing different data sources. Thus, the non-federated multi-task performance of SL is an interesting open question. 

In this work, we systematically studied the multi-task performance of SL and proposed a robust Multi-Task Split Learning (MTSL) framework as an alternative to FL. In this setup, instead of an obstacle, heterogeneity becomes a useful property to take advantage of. Specifically, we showed that when allowing data and computation heterogeneity,  MTSL can have unique advantages in terms of communication cost, convergence rate, and robustness to noise. 

\subsection{Main Contributions}

\begin{itemize}[leftmargin=*]
    \item We propose a robust Multi-Task Split Learning (MTSL) framework for the multi-task scenario, which accommodates data heterogeneity and varying computation capacities among clients. This framework can serve as an alternative to the popular FL framework.
    \item We proved the convergence result of MTSL for a general gradient descent case with convex / non-convex objective functions. Specifically, we showed that MTSL can achieve a fast convergence rate by tuning the learning rate correctly compared with FL. We also proved weaker results for the stochastic gradient descent (SGD) case.
    \item Compared with existing multi-task FL methods, we showed that the MTSL framework has faster convergence, smaller communication cost, and stronger robustness to noise in the multi-task setup through numerical analysis on various image classification datasets.
\end{itemize}

\subsection{Related Works}

As an active research field, there are many interesting works related to multi-task distributed machine learning which we cannot list exhaustively. Here we referenced the four most related research fields: Federated Learning, Split Learning, Split Federated Learning, and Multi-Task Learning.

Federated Learning (FL) trains a full model on the distributed client with their local data and later aggregates the local gradients to update a global model in the server \cite{mcmahan2017communication, kairouz2021advances,li2020review}. FL has the advantage of protecting the privacy of each client while at the same time aggregating the local information through gradient sharing. However, it has been shown that the performance of FL can drop significantly when clients do not have i.i.d data \cite{li2020federated, zhao2018federated, li2022federated, ma2022state}. In addition, the communication cost of transmitting gradients information of large models and the computation cost of hosting a large model on the client side can be significant, especially when the edge device has limited computation and communication capacity \cite{konevcny2016federated, park2021communication, pal2021server}. 

Split Learning (SL) splits the full model into multiple smaller networks and trains them separately on a server and distributed clients with their local data \cite{gupta2018distributed, vepakomma2018split, singh2019detailed}. Instead of the full gradients, each client only uploads its final layer output to the server. In this way, SL reduces the communication cost while protecting the privacy. However, the original sequential SL does not take advantage of parallel computing and has high latency. Later, parallel SL was proposed but its performance was shown to be inferior to FL due to server-client update imbalance \cite{pal2021server, joshi2021splitfed, gao2020end}.  However, the evaluation was done on a common-task setup. The performance of parallel SL on the multi-task setup is unexplored.

Split Federated Learning (SplitFed) are methods that combine SL and FL \cite{thapa2022splitfed, abedi2023fedsl, gupta2018distributed}. Compared with FL, instead of sharing parameters of a full model, clients only have part of the model. Therefore, the communication cost is reduced. Compared with SL, the federation process makes the model perform better in the common-task scenario\cite{han2021accelerating, oh2022locfedmix, liao2023accelerating}. However, in the multi-task scenario, the performance of SplitFed and the necessity of the federation process have not been systematically studied. 

Multi-Task Learning (MTL)  \cite{caruana1997multitask, smith2017federated, zhang2018overview} aims to learn a model that can solve multiple related tasks simultaneously. In the centralized case, multi-task learning usually trains a common large model first and fine-tunes the common model over multiple tasks \cite{sener2018multi, fifty2021efficiently, crawshaw2020multi}. 
In the distributed setup, multi-task FL was studied under various scenarios. For example, using linear models \cite{smith2017federated}, linear combination of pre-trained models \cite{jiang2019improving}, hyper-network communication \cite{shamsian2021personalized}, and mixture of distributions \cite{marfoq2021federated}, etc. These assumptions make the algorithms more analyzable. However, these methods are still based on the federated setup that relies on explicit parameter sharing between clients and a server. Therefore, they still have the disadvantage of FL in terms of communication and computation costs.

\section{Problem Formulation}\label{sec:problem}

Consider an edge network with a common server and $M$ clients. Each client $m$ has its local data distribution $D_m$ over $\mathcal{X}\times \mathcal{Y}$ for its own task. The distributions $\{D_m\}_{m=1}^M$ are in general different but can have some similarity. Consider the learned model for task $m$ as $F_m(\btheta_m, \cdot)$  with parameter $\btheta_m$. Then the empirical estimation of input $X_m$ can be written as
\begin{align}
    \hat{Y}_m = F_m(\btheta_m, X_m)
\end{align}
In the Multi-Task Learning (MTL) setup, the objective is to minimize the loss of all tasks.
\begin{align}
    \min \bE_{D_1, \ldots, D_M}\Big[\sum_{m=1}^M L(Y_m, \hat{Y}_m) \Big]
\end{align}
where $L(\cdot, \cdot)$ is the loss function, $Y_m$  is the true label for input $X_m$. For notation simplicity, we assume each data source has a similar amount of data. If not, weights can be assigned to the losses of each task. For example, based on the data size, we can assign weight $\delta_m = \frac{|D_m|}{\sum_{i=1}^M|D_i|}$.

If all data sources are i.i.d, then setting $F_m$ to be the same for all tasks can be an effective method. For example, in Federated Learning, all clients share the same model and send gradients to the server; the server then aggregates and shares a common set of parameters with the clients. 

In the case of heterogeneous data sources and clients, the FL framework has three potential drawbacks: 
(1) The federation process can hurt the performance of the MTL because gradient information for different tasks can conflict with each other. This will result in slower convergence and/or worse performance of the MTL objective. (2) Due to heterogeneous edge device conditions, some clients may not have the computation power to run the whole model. (3) Transmitting $|\btheta|$ number of parameters and gradients can incur high communication costs.

\begin{figure}[htbp]
\centerline{\includegraphics[width=0.8\linewidth]{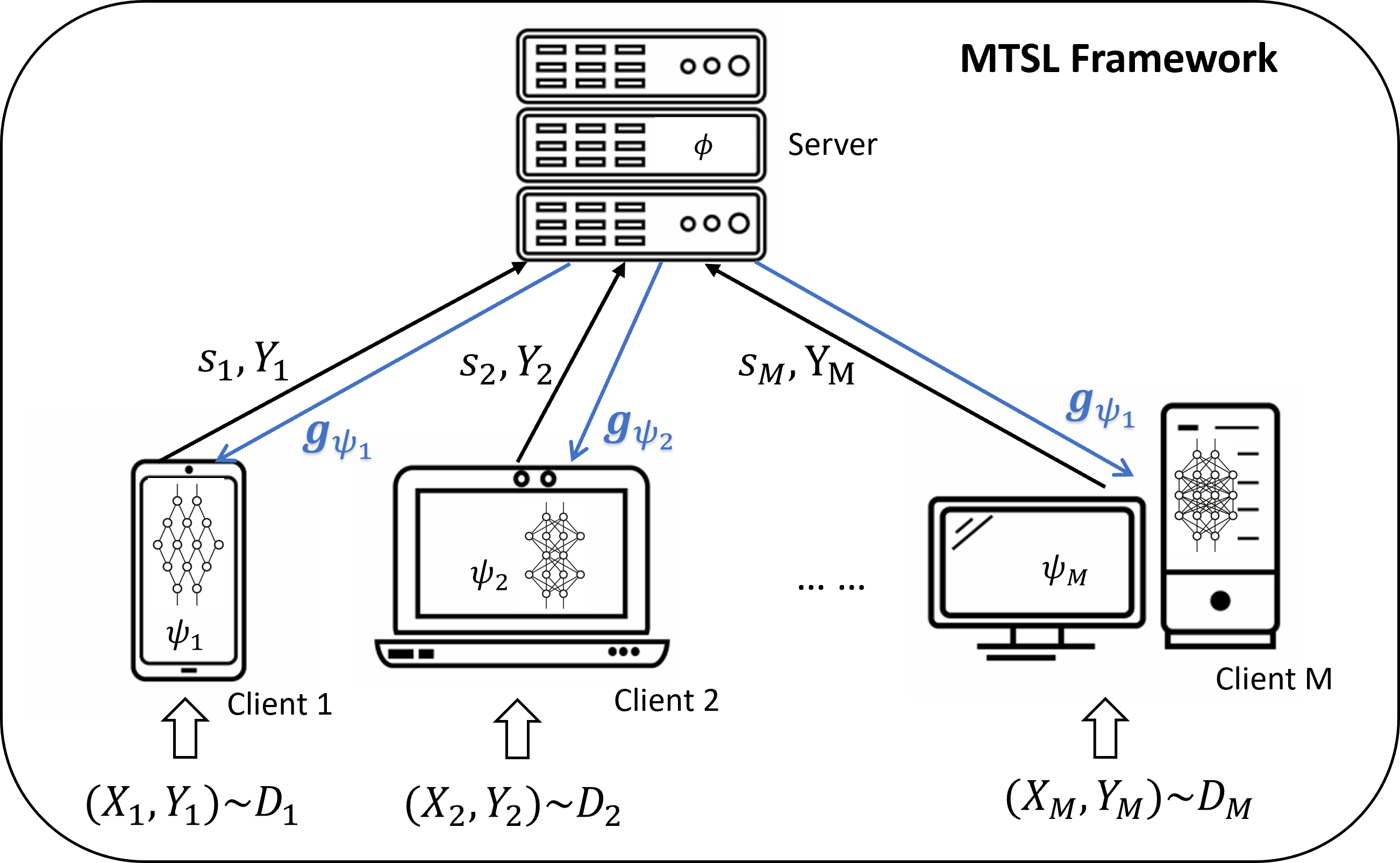}}
\caption{Multi-Task Split Learning Framework. Each client only uploads its smashed data $s_m$ and label $Y_m$ to the server. The server calculates the loss and does the backpropagation of the split network to each client.}
\label{fig:mtsl}
\end{figure}

\subsection{Multi-Task Split Learning (MTSL) Framework}

To address the drawbacks of the FL framework for heterogeneous MTL objective, we propose an alternative framework: Multi-Task Split Learning (MTSL). The schematics are shown in Figure~\ref{fig:mtsl}.

First, to accommodate for the MTL objective, each task can have its own model $F_m$. Second, inspired by the Split Learning case, each model $F_m$ is split between a common server and client $m$. Specifically, for a layered deep neural network $F_m(\btheta_m, \cdot)$, we can split it between a common server and clients,
\begin{align}
    F_m(\btheta_m, \cdot) = G(\bphi,H_m(\bpsi_m,\cdot))
\end{align}
where $G(\bphi,\cdot)$ is the server model with parameter $\bphi$ and $H(\bpsi_m,\cdot)$ is the model of client $m$ with parameter $\bpsi_m$. Client $m$ sends its output (smashed data) $s_m$ and the corresponding labels $Y_m$ to the server and receives the backpropagation gradient results $\bg_{\bphi_m}$ from the common server.

\textbf{Notations}: Vectors are denoted in bold letters. $\odot$ is the element-wise product.  $\lVert\cdot\rVert$ denotes the $l_2$ norm of vectors. $\bg$ and $\Tilde{\bg}$ denote the gradients and gradients estimation respectively. For simplicity, we denote the parameter for the model of task $m$ as $\btheta_m = (\bphi, \bpsi_m)$ and denote the parameter of all models (server and clients) as $\btheta = (\bphi, \bpsi_1, \ldots, \bpsi_M)$. The lower case function denotes the expected function of the objective. Specifically, for task $m$, 
\begin{align}
    f_m(\btheta_m) &= \bE_{D_m} \big[L(Y_m, F_m(\btheta_m, X_m) \big] = \bE_{D_m} \big[L(Y_m, G(\bphi,H_m(\bpsi_m,X_m) \big]
\end{align}

Algorithm~\ref{alg:mtsl} shows the detailed operation schema of the MTSL framework. Note that the server and clients can have different learning rates and model parameters. This gives great flexibility for the MTSL framework to adapt to heterogeneous data sources and clients. 

\begin{algorithm}[H]
\caption{Multi-Task Split Learning Framework (MTSL)}\label{alg:mtsl}
\begin{algorithmic}[1]
\State \textbf{Input:} Data $\{D_m\}_{m=1}^M$, Learning Rates $\eta_s, \{\eta_m\}_{m=1}^M$
\For {iterations $t= 1, 2, \ldots$}
 \State /* Run on Clients */
 \For {each client $m$ in parallel}
    \State Smashed data $s_m = H_m(\bpsi_m, X_m)$
    \State Upload $(s_m, Y_m)$ to the server
 \EndFor

\State /* Run on Server*/
\State Calculate $\hat{Y}_m = G(\bphi, s_m)$ and the loss $L(Y_m, \hat{Y}_m)$ for data from all clients.

\State Run backpropagation and get the gradients for server $\bg_{\bphi}$ and each client $\{\bg_{\bpsi_m}\}_{m=1}^M$

\State Update server parameter $\bphi = \bphi - \eta_s \bg_{\bphi}$

\State /*Run on Clients*/
\For {each client $m$ in parallel}
    \State Download $\bg_{\psi_m}$
    \State Update client parameter $\bpsi_m = \bpsi_m - \eta_m \bg_{\psi_m}$
 \EndFor

\EndFor
\end{algorithmic} 
\end{algorithm}

Compared with the FL framework, MTSL does not have the explicit federation process of gradients aggregation and parameter sharing. Instead, MTSL uses the common server as an implicit way to extract common information from different clients. As we will see in the analysis and simulation later, the MTSL framework can have unique advantages when the data heterogeneity is high.
In addition, similar to split learning, the MTSL framework only transmits the smashed data and gradients of part of the model. Therefore, the communication cost can be reduced.

One potential concern for the MTSL framework is privacy concern because the label needs to be transmitted to the server for each data point, while in the FL framework,  only gradients and parameters are transmitted. This can be addressed by incorporating more complex structures like the U-shape split learning \cite{vepakomma2018split} and privacy protection algorithms like differential privacy \cite{dwork2014algorithmic, dwork2006calibrating}. Since this paper mainly focuses on evaluating the MTL performance of the new framework, we will leave the privacy updates as future works.

\section{Convergence Analysis}

In this section, we will analyze the convergence behavior of the MTSL framework. First, we studied the non-stochastic case, in which we show that heterogeneity can be taken into account by tuning the learning rate accordingly. Then we generalize to the Stochastic Gradient Descent (SGD) case, in which we show a weaker convergence result.
As in the literature, we make some common assumptions:

\begin{assumption}[Lipschitz Continuous Gradient]\label{assmp:l_gradient}
The functions $\{f_m\}_{m=1}^M$ are differentiable and the gradients are Lipshitz continuous with $\lVert \nabla f_m(x_1)-\nabla f_m(x_2) \rVert \leq L_m \lVert x_1-x_2 \rVert$. 
\end{assumption}

\begin{assumption}[Bounded Gradients and Variance]\label{assmp:b_gradient}
    The gradients of the functions $\{f_m\}_{m=1}^M$ are bounded by $B > 0$ and have bounded variance  $\sigma_m^2 \leq G$ for all $m$. 
\end{assumption}

For notation simplicity, we denote the whole MTSL model for both server and clients as
$f(\btheta) = \sum_{m=1}^M f_m(\bphi, \bpsi_m)$.
Let $\bdeta = (\eta_s, \eta_1, \ldots, \eta_M)^\intercal$ be the learning rates and $\bL = (L_s, L_1, \ldots, L_M)^\intercal$ be the Lipschitz constants for the server an $M$ clients respectively. First, we give the convergence results of the MTSL framework in the non-stochastic case.

\begin{proposition}[Non-Stochastic Case] \label{prop:non-stochastic}
Under Assumptions~\ref{assmp:l_gradient}-\ref{assmp:b_gradient}, using gradient descent as the optimization method with learning rate $\bdeta = (\eta_s, \eta_1, \ldots, \eta_M)^{\intercal}$ satisfying $\eta_m \leq \frac{1}{L_m}, \forall m$ , the MTSL framework has the follwing convergence results:

\begin{itemize}[leftmargin=*]
     \item If $f(\btheta)$ is convex, then after $T$ rounds of iterations, the optimality gap satisfies
\begin{align}
    f(\btheta(T)) - f(\btheta^*) =  O\Bigg(\frac{\lVert \frac{1}{\sqrt{\bdeta}} \odot(\btheta(0) - \btheta^*)\rVert^2}{T} \Bigg)
\end{align}
where $\btheta(0)$ is the initial parameter value and $\odot$ is the element wise product.

\item If $f(\btheta)$ is non-convex, then after $T$ rounds of iterations, it converges toward a stationary point with
\begin{align}
    \min_{t \in 1,\ldots,T} \lVert  \sqrt{\bdeta} \odot \nabla f(\btheta(t))\rVert_2^2 = O\Bigg( \frac{f(\btheta(0))-f(\btheta^*)}{T} \Bigg)
\end{align}
where $\btheta(0)$ is the initial parameter value and $\odot$ is the element wise product.
\end{itemize}
\end{proposition}

As we can see, compared with the FL framework, MTSL has the flexibility to tune the learning rate for different clients and the server. This can potentially lead to faster convergence compared with the federation process where all clients share the same model and learning rate, especially when data sources are heterogeneous.

\begin{figure}[htbp]
\centerline{\includegraphics[width=0.85\linewidth]{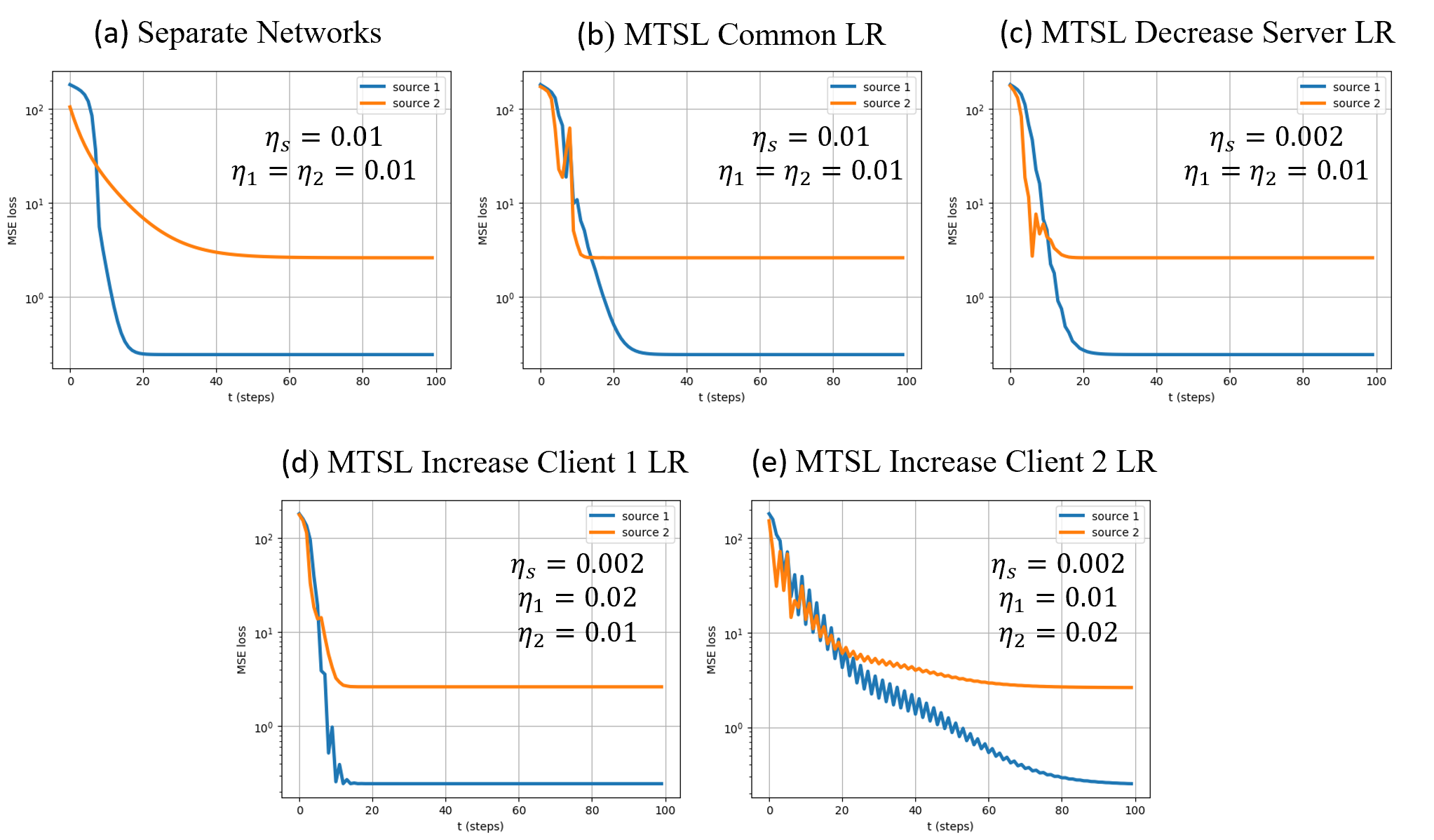}}
\caption{Effect of learning rate tuning for linear model with quadratic loss. (a) Using separate networks for task 1 and 2. (b) MTSL setup with common LR $\eta_s=\eta_1=\eta_2=0.01$. (c) MTSL setup with $\eta_1=\eta_2=0.01$ and decreased server LR $\eta_s=0.002$. (d) MTSL setup with $\eta_s=0.002$, $\eta_2=0.01$ and increased LR for client 1: $\eta_1=0.02$. (e) MTSL setup with $\eta_s=0.002$, $\eta_1=0.01$ and increased LR for client 2: $\eta_1=0.02$.}
\label{fig:linear_quadratic}
\end{figure}
 
To show this effect of LR tuning in MTSL, we consider the special case of linear models with quadratic loss function. Specifically, we consider the model for task $m$ as
\begin{align}
    H_m(X_m) &= b_m X_m + a_m\\
    F_m(X_m) &= G(H_m(X_m)) = w(b_m X_m + a_m) + d
\end{align}
Define the loss function as $L(\hat{Y}_m, Y_m) = (\hat{Y}_m-Y_m)^2$. 

In this case, the Lipschitz constants $\bL=(L_s, L_1, \ldots, L_M)$ satisfy
\begin{align}
    L_s &= \max\{2M, 2\sum_i(b_i^2\bE[X_i^2]+a_i^2)\}\\
    L_i &= \max\{2w^2, 2w^2 \bE[X_i^2]\}, \; i=1,\ldots , M
\end{align}
Therefore, when setting the learning rate (LR), the server LR depends on the number of clients and the sum of the second moment of all clients. On the other hand, client LR depends on the server parameter and its own second moment. The MTSL framework can take advantage of this interdependence and improve the convergence behavior.

To show this behavior, we ran a simulation with the linear model and quadratic loss for different learning rates. Specifically, for a MTSL setup with 2 clients, all models for clients and the server are linear models. The second moment of data source 2 is set to be larger than source 1 with $\bE[X_2^2] = 10\bE[X_1^2]$.

Compared with Figure~\ref{fig:linear_quadratic} (a) which uses a separate network for each task, Figure~\ref{fig:linear_quadratic} (b) shows that without changing the LR, using MTSL setup can help improve the convergence speed for task 2 but not task 1. This may suggest the common LR is too large. Therefore, as shown in Figure~\ref{fig:linear_quadratic} (c), reducing the common LR can improve the convergence speed for both tasks. We can further improve the convergence by doubling the LR for client 1, as shown in Figure~\ref{fig:linear_quadratic} (d). However, doubling the LR for client 2 will hurt the convergence (Figure~\ref{fig:linear_quadratic} (e)). This is because client 2 has a data source with a larger second moment and hence tighter LR range to choose from. 

This relatively simple case of linear models with quadratic loss verified the proposition and the advantage of the MTSL framework. More complex simulations on real datasets will be shown in Section~\ref{sec:simulation}. Next, we will show the convergence results for MTSL in the stochastic gradient case with the unbiased gradient estimation assumption.

\begin{assumption}[Unbiased Gradients Estimation]\label{assmp:unbiased}
  The stochastic estimation of the gradients is unbiased for all clients, i.e. $\bE[\Tilde{\bg}(t) \mid \btheta(t)] = \nabla f(\btheta_t)$.
\end{assumption}

\begin{proposition}[Stochastic Case]\label{prop:stochastic}
 Under Assumptions~\ref{assmp:l_gradient}-\ref{assmp:unbiased}, using the SGD as the optimization method with learning rate $\bdeta = (\eta_s, \eta_1, \ldots, \eta_M)^{\intercal}$, the MTSL framework has the following convergence results:
 \begin{itemize}[leftmargin=*]
     \item If $f(\btheta)$ is convex, then after $T$ rounds of iterations, the optimality gap satisfies
\begin{align}
     \min_{t\in 1\ldots T} \bE[f(\btheta(t)-f(\btheta^*)] = O \Bigg( \frac{ \lVert \btheta(0)- \btheta^* \rVert^2+G^2\sum_{t=1}^T \eta^2_{\min}(t)}{\sum_{t=1}^T \eta_{\min}(t) } \Bigg)
\end{align}
where $\eta_{\min}$ is the minimum element of all learning rates in $\bdeta$.

\item If $f(\btheta)$ is non-convex, then after $T$ rounds of iterations, it converges toward a stationary point with
\begin{align}
    \min_{t\in 1\ldots T} \bE[\lVert \nabla f(\btheta(t)) \rVert^2]  = O\Bigg( \frac{f(\btheta(0))-f(\btheta^*)+ L_{\max}^2B^2 \sum_{t=1}^T \eta_{\max}^2(t)}{\sum_{t=1}^T \eta_{\min}(t)} \Bigg)
\end{align}
where $\eta_{\min}$ is the minimum element of all learning rates in $\bdeta$, $L_{\max}$ is the maximum of all Lipshitz constants defined in Assumption~\ref{assmp:l_gradient}.
 \end{itemize}
\end{proposition}

If the unbiased gradient estimation assumption is not satisfied but instead there is an element-wise bound on the gradient bias, we can still generalize the results in Proposition~\ref{prop:stochastic}
for the convex case. 

\begin{corollary}\label{corolla}
If the bias of the gradient estimation satisfies $    | \bE[\Tilde{\bg}(t)-\nabla f(\btheta_t) \mid \btheta(t)] | \preceq  \xi  |\nabla f(\btheta_t)$, where $\preceq$ means the inequality is satisfied for each element, then the convex results in Proposition~\ref{prop:stochastic} can be modified as $\min_{t \in 1,\ldots,T} \bE[|f(\btheta(t)-f(\btheta^*)|] = O \Big( \frac{ \lVert \btheta(0)- \btheta^* \rVert^2+G^2\sum \eta^2_{\min}(t)}{\sum \eta_{\min}(t) (1-\xi)} \Big)$.
\end{corollary}

Compared with the results in Proposition~\ref{prop:non-stochastic}, the stochastic convergence results are weaker in the sense it only uses the minimum or maximum of the learning rates and Lipschitz constants. Using more sophisticated inequalities can potentially improve this convergence results, which we will leave as a future work. Instead, we will verify the performance of the MTSL framework with SGD using simulations.

\section{Experiments} \label{sec:simulation}

In this section, we will evaluate the performance of the MTSL framework through numerical simulations. Compared with the FL framework, the MTSL framework does not have the explicit federation process which aggregates the gradient information and shares the common parameters across clients. In most existing works for distributed multi-task learning with clients and the server, the federation process has been treated as a necessary step to ensure good performance. Therefore, our main objective in the simulation is to verify that the MTSL framework can perform well when data sources are heterogeneous. In addition to performance, we will also evaluate the cost and robustness of different frameworks.

\subsection{Experiment Setup}

\textbf{Datasets and Models}. We evaluated the multi-task performance of different frameworks on four benchmark image classification datasets. Fashion-MNIST \cite{xiao2017fashion}, EMNIST\cite{cohen2017emnist},   CIFAR10, and CIFAR100 \cite{alex2009learning}. For each task, we use one class (or superclass in CIFAR-100) whose label we denoted as the main label of that task and randomly select samples from the other classes with probability $\alpha$. Specifically, for a dataset with $M$ classes (tasks), the label distribution for task $m$ satisfies,
\begin{align}
    \bP(Y_m=m) = 1-\alpha; \quad \bP(Y_m=n) = \frac{\alpha}{M-1}, \forall n\neq m.
\end{align}

Therefore, the degree of heterogeneity is controlled by $\alpha \in [0, 1-\frac{1}{M}]$. When $\alpha=0$, each task only contains one class, representing the maximum heterogeneity. When $\alpha=1-\frac{1}{M}$, all tasks contain i.i.d. distribution from all classes. We used the given training set for each dataset for training and test on the testing set (10,000 samples).

For MNIST and Fashion-MNIST datasets, we used a 4-layer Multi-Layer Perceptron (MLP) by transforming the original image into a vector directly without using convolution layers. In the MTSL setup, two layers are in clients and 2 layers are in the server. For CIFAR datasets, we used Resnet-16 as the total model. In the MTSL setup,  we split 9 layers in the client and 7 layers in the server.

\begin{table}[htbp]
  \caption{Datasets and Models}
  \label{table:dataset}
  \centering
  \begin{tabular}{>{\centering\arraybackslash}p{0.2\textwidth}|>{\centering\arraybackslash}p{0.2\textwidth}|>{\centering\arraybackslash}p{0.15\textwidth}|>{\centering\arraybackslash}p{0.35\textwidth}}
    \toprule
    Dataset     & Number of Classes     & Total Samples    & Models    \\
    \midrule
    MNIST      & 10      & 70,000      &  Multi-Layer Perceptron (MLP)      \\
    Fashion-MNIST      & 10      & 70,000      &  Multi-Layer Perceptron (MLP)     \\
    CIFAR10     & 10      & 60,000      &  Resnet-16          \\
    CIFAR100    & 10 superclass      & 60,000      &  Resnet-16        \\
    \bottomrule
  \end{tabular}
\end{table}

\textbf{Baseline Algorithms.} We consider three baseline algorithms to compare to: (1) FedAvg\cite{mcmahan2017communication}, the classic federated learning algorithms that first proposed the federation process. (2) FedEM\cite{marfoq2021federated}, a multi-task modification of the FL framework that uses a mixture of distributions to try to address data source heterogeneity. (3) SplitFed \cite{thapa2022splitfed}, a framework that combines FL and SL. instead of the whole model, the federation process is only done for the split-part in clients.

\textbf{Evaluation Methods.} To evaluate the Multi-Task Learning performance of the algorithms, for each task, we will only test on the main label of that task. Therefore, we can view samples from other classes as noise which is controlled by the heterogeneity sampling parameter $\alpha$. In addition, we also allow adding pixel-wise random Gaussian noise. We use the average test accuracies over all tasks to measure the performance. Specifically,
\begin{align}
    {\tt Accuracy_{MTL}} = \frac{1}{M} \sum_{m=1}^M \frac{\text{ Number of Correct Predictions for Task m}}{\text{Number of Test Samples for Task m}}
\end{align}

To test the robustness of the algorithms, we test the performance over different levels of data heterogeneity and noise level. In addition, we also tested the continuous training ability by leaving one client out in the first phase of the training, and added the client back in the second phase of the training without changing the parameters of the other parts of the models for the MTSL framework.

In addition to MTL accuracy and robustness, we also compared: (i) The training cost in terms of model complexity and training steps/time; (ii) The network traffic in terms of the amount of data transmitted between clients and the server.

\subsection{Results}

\textbf{Multi-Task Performance.} Table~\ref{table:accuracy} shows the multi-task testing accuracy for different algorithms when the data sources have high heterogeneity ($\alpha=0$). Compared with the FL-based algorithms, the MTSL framework has higher accuracy across different datasets. The reason is that FL-based algorithms (even the multi-task version like FedEM) use the federation process which cannot deal with conflicting gradient information under heterogeneous data sources. MTSL framework, on the other hand, has a client model for each data source and aggregates the information implicitly using the server model. Therefore, the MTSL framework outperforms FL-based algorithms drastically.

\begin{table}[htbp]
  \caption{Multi-Task Test accuracy of different Algorithms}
  \label{table:accuracy}
  \centering
  \begin{tabular}{>{\centering\arraybackslash}p{0.2\textwidth}|>{\centering\arraybackslash}p{0.1\textwidth}|>{\centering\arraybackslash}p{0.1\textwidth}|>{\centering\arraybackslash}p{0.1\textwidth}|>{\centering\arraybackslash}p{0.15\textwidth}}
    \toprule
    Dataset     & FedAvg     & FedEM    & SplitFed   &MTSL (Ours) \\
    \midrule
    MNIST      & 79.5      & 81.2      &  79.8        &96.8       \\
    Fashion-MNIST      & 78.5      & 79.9      &  78.8        &94.8       \\
    CIFAR10     & 68.2      & 78.6      &  74.5        &92.4       \\
    CIFAR100    & 46.7      & 55.2      &  51.3             &60.2       \\
    \bottomrule
  \end{tabular}
\end{table}

\textbf{Adding a New Client.} We tested the setup where one client was left out in the first phase of training and was only added back in the second phase. The data for the left-out client was also excluded from the first phase of training. When adding the new client, for FL-based algorithms, the federation process is still used so all the other clients are trained at the same time; for the MTSL framework, only the new client model is trained while the models for the other clients are frozen. 

Table~\ref{table:unseen_clients} shows the multi-task testing accuracy for different algorithms after adding new client with unseen training data. The data sources have high heterogeneity ($\alpha=0$). As can be expected, in general, there is a slight drop in performance compared with Table~\ref{table:accuracy} across different algorithms. However, the MTSL algorithm still outperforms the FL based algorithms drastically. In addition, since the MTSL framework only trains the new client, the training cost is less than its FL counterpart.

\begin{table}[htbp]
  \caption{Multi-Task Test accuracy of different Algorithms with unseen clients}
  \label{table:unseen_clients}
  \centering
  \begin{tabular}{>{\centering\arraybackslash}p{0.2\textwidth}|>{\centering\arraybackslash}p{0.1\textwidth}|>{\centering\arraybackslash}p{0.1\textwidth}|>{\centering\arraybackslash}p{0.1\textwidth}|>{\centering\arraybackslash}p{0.15\textwidth}}
    \toprule
    Dataset     & FedAvg     & FedEM    & SplitFed   &MTSL (Ours) \\
    \midrule
    MNIST      & 77.4      & 80.3      &  78.6        &95.4       \\
    Fashion-MNIST      & 76.3      & 77.3      &  76.4        &93.3       \\
    CIFAR10     & 67.1      & 76.9      &  75.3        &91.5       \\
    CIFAR100    & 45.2      & 54.2      &  50.1             &58.1       \\
    \bottomrule
  \end{tabular}
\end{table}

\begin{figure}[htbp]
\centerline{\includegraphics[width=0.8\linewidth]{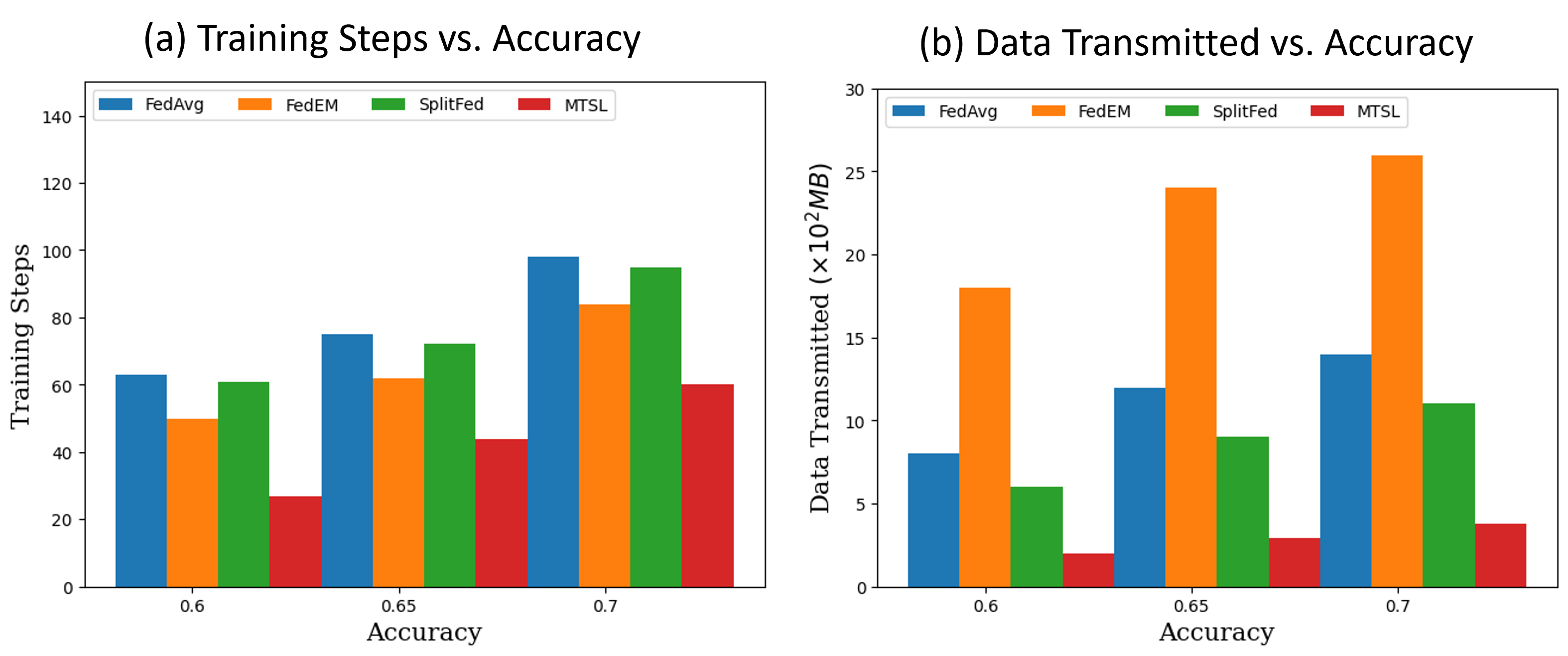}}
\caption{Training cost of different algorithms for MNIST dataset as the testing accuracy increases ($\alpha=0$). (a) Number of training steps needed to reach certain accuracy. (b) Amount of data (smashed data, gradients, parameters) transmitted to reach certain accuracy.}
\label{fig:training_cost}
\end{figure}

\textbf{Training Cost.} To measure the training cost, we recorded the number of training steps and data transmitted when reaching a certain level of accuracy for each algorithm. Figure~\ref{fig:training_cost} shows the results of different algorithms for the MNIST dataset when data sources have high heterogeneity ($\alpha=0$). As we can see in  Figure~\ref{fig:training_cost}(a), the MTSL framework takes fewer training steps to reach the same level of accuracy. This corroborates the propositions we have that the MTSL framework can speed up the training. Figure~\ref{fig:training_cost}(b) shows that the MTSL framework saves the most in transmitted data when dealing with heterogeneous data sources. There are two reasons for this: First, compared with FedAvg and FedEM, MTSL is a split learning based algorithm that only transmits smashed data and client parameters. A similar reduction can be seen for the SplitFed algorithm. Second, MTSL converges faster for heterogeneous tasks and does not use the federation process. So the transmitted data is even smaller than that of SplitFed. 

\begin{figure}[htbp]
\centerline{\includegraphics[width=0.8\linewidth]{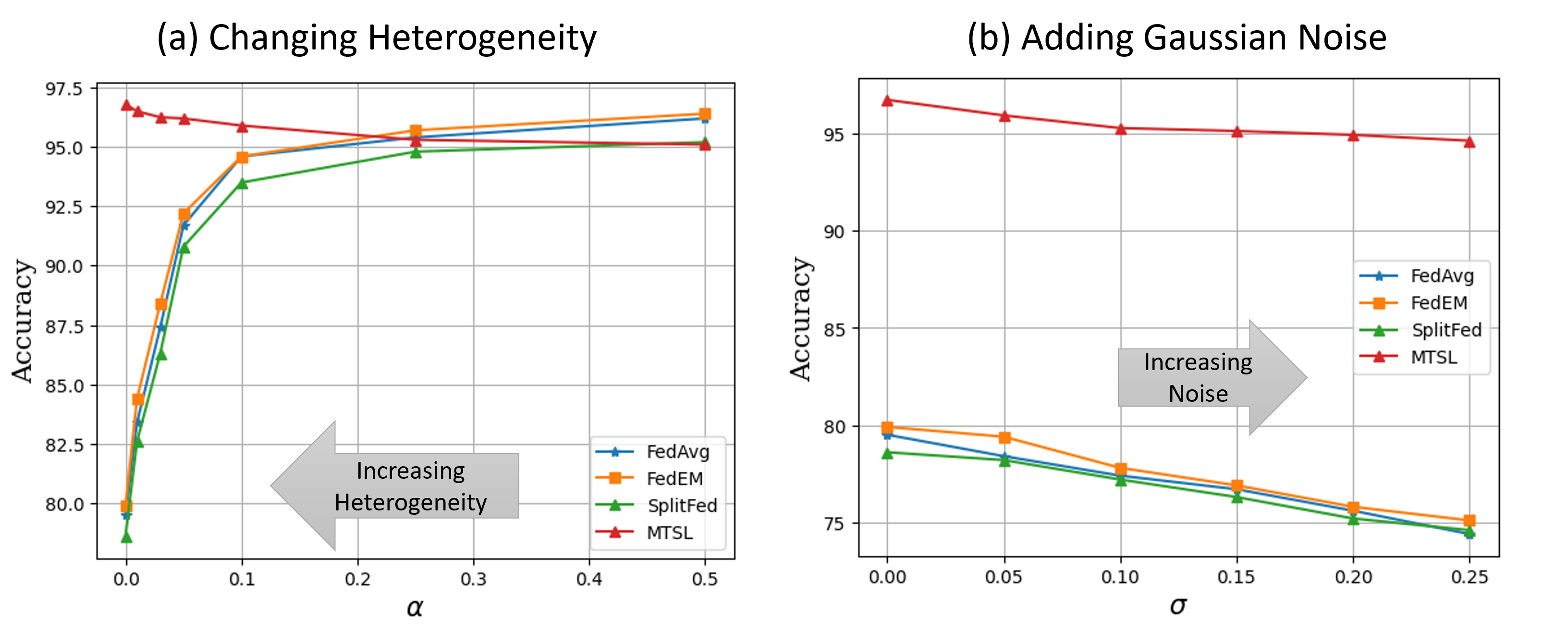}}
\caption{Performance of different algorithms for MNIST dataset as the noise level changes. (a) Changing the data heterogeneity parameter $\alpha$. (b) Adding pixel-wise zero mean Gaussian noise with different standard deviation $\sigma$ ($\alpha=0)$. }
\label{fig:adding_noise}
\end{figure}

\textbf{Robustness to Noise.} Finally, we tested the robustness of different algorithms when changing the data heterogeneity and noise level. Figure~\ref{fig:adding_noise} shows the results for the MNIST dataset for different values of heterogeneity parameter $\alpha$ and Gaussian noise standard deviation $\sigma$. As we can see in Figure~\ref{fig:adding_noise}(a), the performance of the MTSL framework is comparable to other for homogeneous scenario ($\alpha\approx 0.5$) and becomes stable as the data heterogeneity increases ($\alpha \approx 0$). The FL-based algorithms, however, have a sharp performance drop when data heterogeneity becomes larger ($\alpha \approx 0$). This further verifies the claim that the federation process is not an effective method to deal with data heterogeneity. On the other hand, if the data sources shift toward a more i.i.d. distribution ($\alpha\approx 0.5$), the FL-based algorithms show a slightly better but comparable performance to our MTSL-based algorithm. When adding pixel-wise zero mean Gaussian noise (Figure~\ref{fig:adding_noise} (b)), we see a general drop in performance across different algorithms, but the MTSL framework still performs the best compared with the FL-based algorithms. 

\section{Conclusions} \label{sec:conclusion}

In this paper, we proposed a Multi-Task Split Learning (MTSL) framework for distributed multi-task learning with clients and a server. Compared with the classic federated setup, the MTSL framework does not have the explicit federation process which aggregates the gradients and shares parameters of a common model with all clients. Instead, the new framework has more flexibility in setting up the learning rate and hence can have faster convergence. In addition, MTSL is more robust to data heterogeneity and computation power heterogeneity among clients. The claims are verified through numerical studies for multiple datasets. 

It is worth noting that the MTSL framework is not meant to replace the FL framework. As we have shown in the experiment results, FL still has advantages when data distribution shifts toward i.i.d. Therefore, as an alternative framework, MTSL is more suitable to handle scenarios where data sources and clients have high heterogeneity, as can be expected in future network systems. With these in mind, there are still many open questions for the MTSL framework, including how to dynamically adapt the MTSL framework to data source heterogeneity and how to improve data privacy.


\begin{ack}
Use unnumbered first level headings for the acknowledgments. All acknowledgments
go at the end of the paper before the list of references. Moreover, you are required to declare
funding (financial activities supporting the submitted work) and competing interests (related financial activities outside the submitted work).
More information about this disclosure can be found at: \url{https://neurips.cc/Conferences/2022/PaperInformation/FundingDisclosure}.

Do {\bf not} include this section in the anonymized submission, only in the final paper. You can use the \texttt{ack} environment provided in the style file to autmoatically hide this section in the anonymized submission.
\end{ack}


\newpage
\bibliography{main}
\bibliographystyle{unsrtnat}

\newpage

\appendix

\section{Appendix}

\subsection{More Related Works}

Here we list more related works on the Federate Learning (FL) framework that are intended to deal with data heterogeneity.

For data-based methods, there are approaches that tried to improve the performance by data sharing \cite{zhao2018federated, tuor2021overcoming} and data augmentation \cite{duan2019astraea, shin2020xor}. However, these approaches require sharing original data sources between clients, which contradicts the original privacy objective of the FL framework.

For model-based methods, there are approaches that tried to adapt the client model by using local fine-tuning \cite{fallah2020personalized, jiang2019improving, hanzely2020federated}, adding personalized layer\cite{arivazhagan2019federated, mansour2020three, liang2020think}, and knowledge distillation\cite{wang2019federated, chang2019cronus, peng2019federated}, client clustering \cite{ghosh2020efficient, briggs2020federated, sattler2020clustered} etc. However, most of these approaches still keep the federation process similar to the FedEM and SplitFed methods we have discussed in the main paper.

\subsection{Proofs}

By definition in Section~\ref{sec:problem},
\begin{align*}
    f(\btheta) &= \bE_{D_1, \ldots, D_M}\Big[\sum_{i=1}^M L(D_i, F_i(\btheta_i,D_i) \Big]\\
    &= \bE_{D_1, \ldots, D_M}\Big[\sum_{i=1}^M L(D_i, F_i(\bphi, \bpsi_i, D_i) \Big]\\
    &= \sum_{i=1}^M f_i(\bphi, \bpsi_i)
\end{align*}

Therefore, the gradient vector can be written as
\begin{align*}
    \nabla f(\btheta) = \big[\sum_{i=1}^M \frac{\partial f_i}{\partial \bphi}, \frac{\partial f_1}{\partial\bpsi_1}, \ldots, \frac{\partial f_M}{\partial\bpsi_M} \big]^\intercal
\end{align*}

\textbf{Proof for Proposition 1.}
Let $\bdeta = [\eta_0, \eta_1, \ldots, \eta_M]^\intercal$ be the learning rate for each component and $\bL = [L_0, L_1, \ldots, L_M]^\intercal$ be the Lipschitz constant of each component. At epoch $t$, using descent lemma,

\begin{align*}
    f(\btheta(t+1)) &\leq f(\btheta(t)) + \nabla  f(\btheta(t))^\intercal  (\btheta(t+1)-\btheta(t)) +  \frac{1}{2}\lVert \bL\odot (\btheta(t+1)-\btheta(t))\rVert_2^2 \\
    & = f(\btheta(t))  - \nabla  f(\btheta(t))^\intercal (\bdeta \odot \nabla  f(\btheta(t))) +  \frac{1}{2} \lVert \bL \odot \bdeta \odot \nabla f(\btheta(t))\rVert_2^2
\end{align*}

If $\eta_i \leq \frac{1}{L_i}$ for each component $i=0,\ldots M$, then
\begin{align*}
    f(\btheta(t+1)) &\leq f(\btheta(t))  - \frac{1}{2}\lVert  \sqrt{\bdeta} \odot \nabla f(\btheta(t))\rVert_2^2
\end{align*}

\textbf{If $f(\btheta)$ is convex, then}
\begin{align*}
    f(\btheta(t)) \leq f(\btheta^*) +  \nabla  f(\btheta(t))^\intercal (\btheta(t) - \btheta^*)
\end{align*}
Then, 
\begin{align*}
    f(\btheta(t+1)) - f(\btheta^*)  &\leq  \nabla  f(\btheta(t))^\intercal (\btheta(t) - \btheta^*)   - \frac{1}{2}\lVert  \sqrt{\bdeta} \odot \nabla f(\btheta(t))\rVert_2^2 \\
    &\leq -\frac{1}{2}\Big(\lVert  \sqrt{\bdeta} \odot \nabla f(\btheta(t))\rVert_2^2 - 2\nabla  f(\btheta(t))^\intercal (\btheta(t) - \btheta^*) + \lVert \frac{1}{\sqrt{\bdeta}} \odot(\btheta(t) - \btheta^*) \rVert_2^2 \\
    &- \lVert \frac{1}{\sqrt{\bdeta}} \odot(\btheta(t) - \btheta^*)\rVert_2^2\Big)\\
    &=- \frac{1}{2}\Big(\lVert \frac{1}{\sqrt{\bdeta}} \odot(\btheta(t+1) - \btheta^*)\rVert_2^2 - \lVert \frac{1}{\sqrt{\bdeta}} \odot(\btheta(t) - \btheta^*)\rVert_2^2 \Big)
\end{align*}
Taking telescoping sum from $t=0$ to $T-1$,
\begin{align*}
    \sum_{t=0}^{T-1} ( f(\btheta(t+1)) - f(\btheta^*)) \leq \frac{1}{2} \lVert \frac{1}{\sqrt{\bdeta}} \odot(\btheta(0) - \btheta^*)\rVert_2^2
\end{align*}
Since $f(\btheta(t))$ is monotone non-increasing, we have
\begin{align*}
    f(\btheta(t)) - f(\btheta^*) \leq \frac{1}{T} \sum_{t=0}^{T-1} ( f(\btheta(t+1)) - f(\btheta^*)) \leq \frac{\frac{1}{2}\lVert \frac{1}{\sqrt{\bdeta}} \odot(\btheta(0) - \btheta^*)\rVert_2^2}{T}
\end{align*}

\textbf{If $f(\btheta)$ is non-convex, then}

\begin{align*}
    f(\btheta(t)) - f(\btheta(0)) &\leq -\frac{1}{2}\sum_{t=0}^{T-1} \lVert  \sqrt{\bdeta} \odot \nabla f(\btheta(t))\rVert_2^2\\
    &\leq -\frac{T}{2} \min_{t \in 1,\ldots,T} \lVert  \sqrt{\bdeta} \odot \nabla f(\btheta(t))\rVert_2^2
\end{align*}

Therefore,
\begin{align*}
    \min_{t \in 1,\ldots,T} \lVert  \sqrt{\bdeta} \odot \nabla f(\btheta(t))\rVert_2^2 \leq \frac{2(f(\btheta(0))-f(\btheta^*))}{T}
\end{align*}

\textbf{Proof for Proposition 2.}
If we have an unbiased estimation of gradients, $\bE[\Tilde{\bg}(t) \mid \btheta(t)] = \nabla f(\btheta_t)$, then
\begin{align*}
   &\bE [ \lVert \btheta(t+1)- \btheta^* \rVert_2^2 \mid \btheta(t)]\\
   & = \bE [ \lVert \btheta(t)- \bdeta(t) \odot \Tilde{\bg}(t) - \theta^* \rVert_2^2 \mid \btheta(t)]\\
   & = \bE [ \lVert \btheta(t)- \btheta^* \rVert_2^2 + \lVert \bdeta(t) \odot \Tilde{\bg}(t) \rVert_2^2 - 2 \bdeta(t) \odot \Tilde{\bg}(t)^\intercal (\btheta(t)- \btheta^*) \mid \btheta(t)]\\
   & = \lVert \btheta(t)- \btheta^* \rVert_2^2 + \bE[\lVert \bdeta(t) \odot \Tilde{\bg}(t) \rVert_2^2 \mid \btheta(t)] - 2 \bdeta(t) \odot \bE[\Tilde{\bg}(t)  \mid \btheta(t)]^\intercal (\btheta(t)- \btheta^*)
\end{align*}

\textbf{If $f(\btheta)$ is convex, then}
\begin{align*}
    f(\btheta^*)\geq f(\btheta(t)) + \bE[\Tilde{\bg}(t)  \mid \btheta(t)]^\intercal (\btheta^* - \btheta(t))
\end{align*}
We have
\begin{align*}
    - 2 \bdeta \odot \bE[\Tilde{\bg}(t)  \mid \btheta(t)]^\intercal (\btheta(t)- \btheta^*) \leq -\eta_{\min}(t)(f(\btheta(t)-f(\btheta^*))
\end{align*}

Taking expectation over the joint distribution of the history,
\begin{align*}
    \bE [ \lVert \btheta(t+1)- \btheta^* \rVert_2^2] \leq \bE [ \lVert \btheta(t)- \btheta^* \rVert_2^2] - 2\eta_{\min}(t)\bE[f(\btheta(t)-f(\btheta^*)]+\bE[\lVert \bdeta(t) \odot \Tilde{\bg}(t) \rVert_2^2]
\end{align*}

Assume $\bE[\lVert\Tilde{\bg}(t) \rVert_2^2] \leq G^2$, taking telescoping sum, we have
\begin{align*}
    \bE [ \lVert \btheta(T+1)- \btheta^* \rVert_2^2] \leq \bE [ \lVert \btheta(1)- \btheta^* \rVert_2^2]-2\sum_{t=1}^T \eta_{\min}\bE[f(\btheta(t)-f(\btheta^*)] + G^2\eta_{\min}(t)^2
\end{align*}

Using the fact that $\bE[f(\btheta(t)-f(\btheta^*)] \geq \min_{t \in 1,\ldots,T} \bE[f(\btheta(t)-f(\btheta^*)]$, we have
\begin{align*}
     \min_{t \in 1,\ldots,T} \bE[f(\btheta(t)-f(\btheta^*)] \leq \frac{[ \lVert \btheta(0)- \btheta^* \rVert_2^2]+G^2\sum \eta_{\min}(t)^2}{2\sum \eta_{\min}(t) }
\end{align*}

\textbf{If $f(\btheta)$ is non-convex, then}

From descent lemma, 
\begin{align*}
     f(\btheta(t+1)) &\leq f(\btheta(t)) + \nabla  f(\btheta(t))^\intercal  (\btheta(t+1)-\btheta(t)) +  \frac{1}{2}\lVert \bL\odot (\btheta(t+1)-\btheta(t))\rVert_2^2\\
     & \leq f(\btheta(t)) - \nabla  f(\btheta(t))^\intercal \bdeta(t)\odot \Tilde{\bg}(t) + \frac{1}{2}\lVert \bL\odot \bdeta(t)\odot \Tilde{\bg}(t))\rVert_2^2
\end{align*}

Taking expectations with respect to samples for $\Tilde{\bg}(t)$,
\begin{align*}
   \bE[ f(\btheta(t+1))] \leq \bE[ f(\btheta(t))]-\eta_{\min}(t) \bE[\lVert \nabla f(\btheta(t) \rVert^2] + \frac{L_{\max}^2\eta_{\max}^2(t)}{2} \bE[\lVert\Tilde{\bg}(t) \rVert_2^2]
\end{align*}

Assuming $\bE[\lVert\Tilde{\bg}(t) \rVert_2^2]\leq G^2$, taking telescoping sum,
\begin{align*}
    \sum_{t=1}^T \eta_{\min}(t) \bE[\lVert \nabla f(\btheta(t) \rVert^2] \leq f(\btheta(0))-f(\btheta^*)+\frac{L_{\max}^2G^2}{2} \sum_{t=1}^T \eta_{\max}^2(t)
\end{align*}
Therefore,
\begin{align*}
    \min_{t \in 1,\ldots,T} \bE[\lVert \nabla f(\btheta(t) \rVert^2]  \leq \frac{f(\btheta(0))-f(\btheta^*)+ \frac{L_{\max}^2G^2}{2} \sum_{t=1}^T \eta_{\max}^2(t)}{\sum_{t=1}^T \eta_{\min}(t)}
\end{align*}

\textbf{Proof of Corollary 1.} 
If the following assumptions are satisfied,
\begin{align*}
    | \bE[\Tilde{\bg}(t)-\nabla f(\btheta_t) \mid \btheta(t)] | &\preceq  \alpha  |\nabla f(\btheta_t)| \\
    \bE[\lVert\Tilde{\bg}(t) \rVert_2^2] &\leq G^2
\end{align*}

Then, in the convex case,
\begin{align*}
    f(\btheta^*)\geq f(\btheta(t)) + (\bE[\Tilde{\bg}(t)  \mid \btheta(t)]^\intercal - \alpha \nabla f(\btheta_t) )(\btheta^* - \btheta(t))
\end{align*}

We have
\begin{align*}
    \bE [ \lVert \btheta(t+1)- \btheta^* \rVert_2^2] \leq \bE [ \lVert \btheta(t)- \btheta^* \rVert_2^2] - 2\eta_{\min}(t)(1-\alpha)\bE[f(\btheta(t)-f(\btheta^*)]+\bE[\lVert \bdeta(t) \odot \Tilde{\bg}(t) \rVert_2^2]
\end{align*}

Using the fact that $\bE[f(\btheta(t)-f(\btheta^*)] \geq \min_{t \in 1,\ldots,T} \bE[f(\btheta(t)-f(\btheta^*)]$, we have
\begin{align*}
     \min_{t \in 1,\ldots,T} \bE[f(\btheta(t)-f(\btheta^*)] \leq \frac{[ \lVert \btheta(0)- \btheta^* \rVert_2^2]+G^2\sum \eta_{\min}(t)^2}{2\sum \eta_{\min}(t)(1-\alpha) }
\end{align*}

\end{document}